\documentclass[journal,comsoc]{IEEEtran}
\usepackage[T1]{fontenc}
\usepackage{url}
\usepackage{graphicx}
\usepackage{subfigure}
\usepackage{multirow}
\usepackage{float}
\usepackage{cite}
\usepackage{amsmath}
\usepackage[british]{babel}

\usepackage{tikz}
\usepackage{lipsum}

\ifCLASSINFOpdf
\else
\fi
\newcommand{\comment}[1]{}
\usepackage{array}
\usepackage{placeins}
\makeatletter
\newcommand{\thickhline}{%
	\noalign {\ifnum 0=`}\fi \hrule height 1pt
	\futurelet \reserved@a \@xhline
}

\usepackage{booktabs}
\newcolumntype{"}{@{\hskip\tabcolsep\vrule width 1pt\hskip\tabcolsep}}
\makeatother

\usepackage{amsmath}
\usepackage{lipsum}
\usepackage[inline, shortlabels]{enumitem}
\setlist{itemjoin ={,\enspace},itemjoin* = { and\enspace}}

\usepackage[cmintegrals]{newtxmath}

\begin{document}

	%
	\title{Deep Autoregressive Models with Spectral Attention}
	%
	%
	%
	
	\author{Fernando~Moreno-Pino,~Pablo~M.~Olmos,~Antonio~Artés-Rodríguez
	\thanks{F. Moreno-Pino and P. M. Olmos are with the Department of Signal Theory and Communications, Universidad Carlos III de Madrid, Madrid 28911, Spain, and with the Health Research Institute Gregorio Marañón, Madrid 28007, Spain (e-mail: fmoreno@tsc.uc3m.es; olmos@tsc.uc3m.es)}
		\thanks{A. Artés-Rodríguez is with the Department of Signal Theory and Communications, Universidad Carlos III de Madrid, Madrid 28911, Spain, with the Health Research Institute Gregorio Marañón, Madrid 28007, Spain, and  with the CIBERSAM, ISCIII, Madrid 28029, Spain (e-mail:, antonio@tsc.uc3m.es).}}

	%
	%

	\markboth{Preprint - Deep Autoregressive Models with Spectral Attention}%
	{Shell \MakeLowercase{\textit{et al.}}: Frecuency Attentive Autoregressive Model}
	%



	\maketitle

	\begin{abstract}

	
	Time series forecasting is an important problem across many domains, playing a crucial role in multiple real-world applications. In this paper, we propose a forecasting architecture that combines deep autoregressive models with a Spectral Attention (SA) module, which merges global and local frequency domain information in the model's embedded space. 
	By characterizing in the spectral domain the embedding of the time series as occurrences of a random process, our method can identify global trends and seasonality patterns. Two spectral attention models, global and local to the time series,  integrate this information within the forecast and perform spectral filtering to remove time series's noise. The proposed architecture has a number of useful properties: it can be effectively incorporated into well-know forecast architectures, requiring a low number of parameters and producing interpretable results that improve forecasting accuracy. We test the Spectral Attention Autoregressive Model (SAAM) on several well-know forecast datasets, consistently demonstrating that our model compares favorably to state-of-the-art approaches.

	\end{abstract}
	
	\begin{IEEEkeywords}
		 Attention models, deep learning, filtering, global-local contexts, signal processing, spectral domain a\-tten\-tion, time series forecasting
	\end{IEEEkeywords}

	%
	\IEEEpeerreviewmaketitle

	%
	%
	%
	%

	\section{Introduction}
	\label{introduction}
	Time series forecasting, which consists of analyzing historical signals patterns to predict future outcomes, is an important problem with scientific, business, and industrial applications, playing an important role in daily life. 
	Several fields benefit from time series forecasting, such as finance applications of these models for estimating the future movements of stock markets \cite{engle1982autoregressive, ding2015deep, chatigny2020financial ,kim2003financial,sezer2020financial}, climate prediction \cite{ak2015interval, le2020deep, shi2015convolutional}, forecasting of energy consumption and demand \cite{masum2018multi, wang2019review,dudek2021hybrid}, and product demand and supply \cite{simchi2008designing,merkuryeva2019demand,akyuz2017ensemble}, which helps optimize resource allocation, allowing for cost reduction and profit maximization. 
	
	Early approaches to solving time series forecasting pro\-blems rely on statistical models, such as State Space Models (SSMs) \cite{durbin2012time}, exponential smoothing, \cite{hyndman2008forecasting}, \cite{McKenzie_1984} or matrix factorization methods \cite{yu2016temporal},, e. \cite{Hyndman_2011}, which learn information via a matrix that relates different time series. Auto Regressive Integrated Moving Average (ARIMA) \cite{box2008time} has been one of the most popular solutions, it produces its predictions as a weighted sum of past observations, making the model vul\-ne\-ra\-ble to error accumulation. For extra information regarding classic techniques, we refer the readers to	\cite{Box_1968,hamilton1994time,lutkepohl2005new}.
	
	Regardless, all these classic approaches share a number of weaknesses. They make linearity assumptions on the data, which together with its limited scalability, makes them unsuitable for modern large-scale forecasting tasks. Furthermore, they incorporate prior knowledge about time series com\-po\-si\-tion, like trends or seasonality patterns present in the data, which requires manual feature engineering and model design by domain experts in order to achieve good results \cite{harvey1990forecasting}. Moreover, they do not usually share a global set of parameters among the different time series, which implies bad generalization and poor performance aside from single-linear time series prediction.
	
	Deep neural networks \cite{sutskever2014sequence}, \cite{graves2013generating} are an alternative solution for time series forecasting. They are able to model non-linear temporal patterns, easily identify complex structures across time series, efficiently extract higher order features, and allow a data-driven approach that requires little to no human feature engineering \cite{hwang2015recurrent, sutskever2014sequence,graves2013generating, giuliari2021transformer}. 
	
	Recurrent Neural Networks (RNNs) \cite{funahashi1993approximation}, \cite{lai2018modeling} and Long Short-Term Memory (LSTM) \cite{hochreiter1997long} have achieved good results in temporal modeling. DeepAR \cite{salinas2020deepar} set a milestone using a LSTM to perform an embedding latter used by a pro\-ba\-bi\-lis\-tic model to forecast in an encoder-decoder fashion. Recently, attention models \cite{bahdanau2014neural}, \cite{li2019enhancing}  have been used by these recurrent architectures to selectively focus on some segments of the data while making predictions, e.g., in machine translation, only certain words in the input sequence may be relevant for predicting the next word. To do so,	these models use an inductive bias that connects each token in the input through a relevance weighted basis of every other token. 
	

	The idea behind recurrent attention leads to Transformers models \cite{vaswani2017attention}, which have become one of the most popular methods with respect to the problem of time series forecasting. Initially introduced for Natural Language Processing (NLP), Transformers proposed a completely new architecture where a self-attention mechanism is used to process sequences of data.
	Several modifications must be accomplished in order to apply Transformer models for the forecasting of time series. In ConvTrans \cite{li2019enhancing}, most of the  di\-ffi\-cul\-ties associated with Transformers for this specific problem were solved. An alternative to the canonical self-attention of these models was designed to make them aware of the local context via a CNN \cite{lecun1995convolutional}. At the same time, a modification of the attention mechanism to reduce the computational cost of self-attention was also introduced.

	Many models have also tried to join classical approaches with deep learning techniques, as Deep State Space Models for Time Series Forecasting (DSSM) \cite{rangapuram2018deep} or Deep Factors for Forecasting \cite{wang2019deep}. 
	Various attempts to merge signal processing techniques with deep neural networks can also be found in the literature. In \cite{tamkin2020language}, a framework that uses spectral filtering for the problem of NLP was proposed, while \cite{Cao2020SpectralTG} uses the spectral domain to jointly capture inter-series correlations and temporal dependencies.
	
	Nevertheless, deep autoregressive models also present some inconveniences. First, they tend to focus on recent past data to predict the future of the time series, frequently wasting important global information not encapsulated in previous pre\-dic\-tions. Second, as classical time series forecasting methods, they suffer from error accumulation and propagation \cite{cheng2006multistep}, a problem closely related to the previous one. Third, they do not produce interpretable results, neither can we clearly explain how they reach them \cite{castelvecchi2016can}.
	
	In this paper we show that the described problems can be partially alleviated by in\-cor\-po\-ra\-ting signal processing filtering techniques into the autoregressive models that perform the time series forecasting. With respect to the inability of these models to focus on the global context, we can obtain time series' most important trends via frequency domain cha\-rac\-te\-ri\-za\-tion. These trends can be intelligently incorporated during the forecasting, hence making the local context aware of the time series global patterns. Regarding error accumulation and noisy local context, spectral filtering can be applied to decide at every time ins\-tant which frequencies are useful and which can be suppressed, eliminating unwanted components that do not help during forecasting. Finally, these signal processing tools which operate in the spectral domain produce more interpretable internal representations, as it will be proved during the experiments section, making it possible to extract the explainable frequency domain features that are driving the predictions if necessary.

	To integrate previous solutions, we propose a general architecture, the Spectral Attention Autoregressive Model (SAAM). SAAM's modularity allows it to be effectively incorporated into a variety of deep-autoregressive models. This architecture uses two spectral attention models to determine, at every time instant, relevant global patterns as well as removing local context's noise while performing the forecasting. Both operations are performed in the frequency domain of the embedded space.

	 To the best of our knowledge, SAAM is the first deep neural autoregressive model that exploits attention me\-cha\-nisms in the spectral domain. 
	A global-local architecture marries deep neural networks with classic signal processing techniques in this new frequency domain attention framework, incorporating relevant global trends into the forecast and performing spectral filtering to prevent error accumulation.
    Further, the additional complexity due to Spectral Attention is comparable to classic attention models in the temporal domain.

	We perform extensive experiments on both synthetic and real-world time series datasets, showing the effectiveness of the proposed Spectral Attention module, consistently outperforming the base models and rea\-ching state-of-the-art results. Ablation studies further prove the effectiveness of the designed architecture.

	The rest of this paper is organized as follows. Section \ref{preliminaries} states the time series forecasting problem, presents a base ar\-chi\-tec\-tu\-re to which we append the proposed Spectral Atten\-tion module, and provides a cha\-rac\-te\-ri\-za\-tion of the time series in the spectral domain. Section \ref{FAAM} describes our model and Section \ref{experiments} proves its effectiveness, both quantitative and qualitatively. We conclude the paper in Section \ref{conclusion}.

	\section{Preliminaries}
	\label{preliminaries}
	
	In this section, we formally state the problem of time series forecasting and introduce a base architecture that represents the core of most deep learning-based autoregressive models in the state-of-the-art. Also,  a fre\-quen\-cy domain characterization of the time series is proposed.
	
	\subsection{Problem definition}
	\label{problem_definition}
	
	
	Given a set of \textit{N} univariate time series $\{z^{i}_{1:t_0-1}\}_{i=1}^{N} $, where $z_{1: t_0-1}^{i}=(z_{1}^{i}, z_{2}^{i}, \ldots, z_{t_{0}-1}^{i})$, $t_0\in  \mathbb{N}$ is the forecast horizon,
	$\tau \in  \mathbb{N}$ the forecast length, and ${T=t_0 + \tau \in  \mathbb{N}} $ the sequences' total length,  our goal is to model the conditional probability distribution of future trajectories of each time series given the past, namely, to predict the next $\tau$ time steps after the forecast horizon:

		\begin{equation}
	\begin{aligned}
	p\left(z_{t_{0}: t_0 + \tau}^{i} \mid z_{1: t_{0}-1}^{i}, \mathbf{x}_{1: t_0 + \tau}^{i}, \theta\right) =\prod_{t=t_{0}}^{t_0 + \tau} p\left(z_{t}^{i} \mid z_{1: t-1}^{i}, \mathbf{x}_{1: t}^{i}, \theta \right),
	\end{aligned}
	\end{equation}
	where  $\theta$ are the learnable parameters of the model and ${\{\mathbf{x}^{i}_{1:t_0+\tau}\}_{i=1}^{N} \in \mathbb{R}^{C}}$ the associated covariates. These covariates are, together with time series' past observations, the input to our predictive model.
		
	

	\subsection{Base Architecture}
	\label{base_architecture}
	
\begin{figure}[H] 
	\begin{center}
		\centerline{\includegraphics[width=0.5\columnwidth]{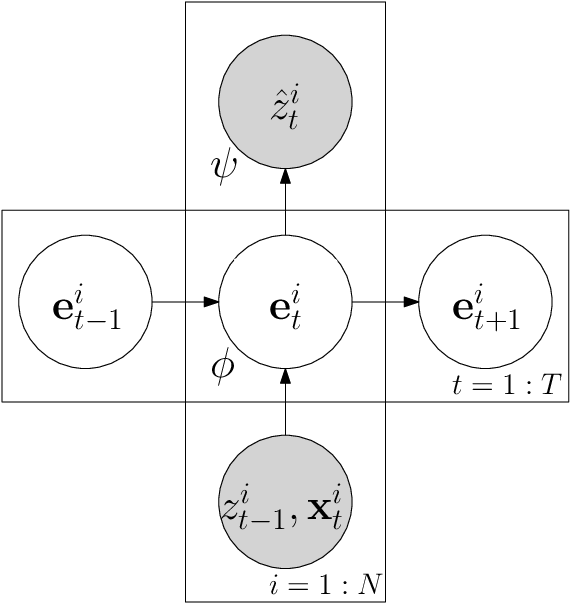}}
		\caption{Common base architecture of deep learning-based autoregressive models. Gray represents observed variables.}
		\label{fig:basic_architecture}
	\end{center}
\end{figure}

	A number of deep-autoregressive models in the state-of-the-art, including DeepAR \cite{salinas2020deepar}, NBeats \cite{oreshkin2019n} and ConvTrans \cite{li2019enhancing}, can be characterized by means of a high-level architecture, represented in Fig. \ref{fig:basic_architecture}. This general framework is composed of two parts:

	\begin{enumerate}
		
				
		
		\item An embedding function $\mathbf{e}_{t}^{i}=f_{\phi}\left(\mathbf{e}_{t-1}^{i}, z^{i}_{t-1}, \mathbf{x}_{t}^{i} \right)  \in \mathbb{R}^{D}$, with transit function $f_{\phi}(\cdot)$ and parameters $\phi$. This em\-be\-dding receives as input, at time $t \in T$, the time series previous value $z_{t-1}^{i}$, the covariates $\mathbf{x}_{t}^{i}$, and the past value of the embedding $\mathbf{e}_{t-1}^{i}$. This embedding function can be implemented in different ways, with a RNN \cite{rumelhart1986learning},\cite{hopfield1982neural}, a LSTM \cite{gers2000learning}, or a Temporal Convolutional Network (TCN) \cite{lea2016temporal}.
		
		\item A probabilistic model  $p\left(z_{t}^{i} \mid  \mathbf{e}_{t}^{i} \right)$, with parameters $\psi$, which uses the embedding $ \mathbf{e}_{t}^{i}$ to estimate time series' next value, $\hat{z}_{t}^{i}$.
		
		This probabilistic model is usually implemented as a function of a neural network that parameterizes the required probability distribution. E.g., a Gaussian distribution can be represented through its mean and standard deviation as: $\mu=g_{\mu}(\mathbf{w}_{\mu}^{T} \mathbf{e}_{t}^{i}+b_{\mu})$, $	\sigma=\log \left(1+\exp \left(g_{\sigma}(\mathbf{w}_{\sigma}^{T} \mathbf{e}_{t}^{i}+b_{\sigma})\right)\right)$, where $g_{\mu}$ and $g_{\sigma}$ are neural networks. 
		
	\end{enumerate}

	The optimal model's parameters $\theta=\{\phi, \psi\}$ are obtained by maximizing the log-likelihood function $\mathcal{L}(\theta)$  for the observed data in the conditioning range, i.e., from $t=1$ to $t_{0}-1$. All quantities required for computing the log-likelihood function are deterministic, which means that no inference is required:

	\begin{equation}
	\begin{aligned}
	&\mathcal{L}(\theta)=\sum_{i=1}^{N}  \log p\left(z_{1:t_0-1}^{i} \mid \mathbf{x}_{1:t_0-1}^{i},\theta\right)  = \\
	&=\sum_{i=1}^{N} \sum_{t=1}^{t_0-1} p\left(z_{t}^{i} \mid \mathbf{x}_{1:t-1}^{i},\theta (\phi, \psi) \right).
	\end{aligned}
	\end{equation}

\begin{figure}[H] 
		\begin{center}
			\centerline{\includegraphics[width=1.0\columnwidth]{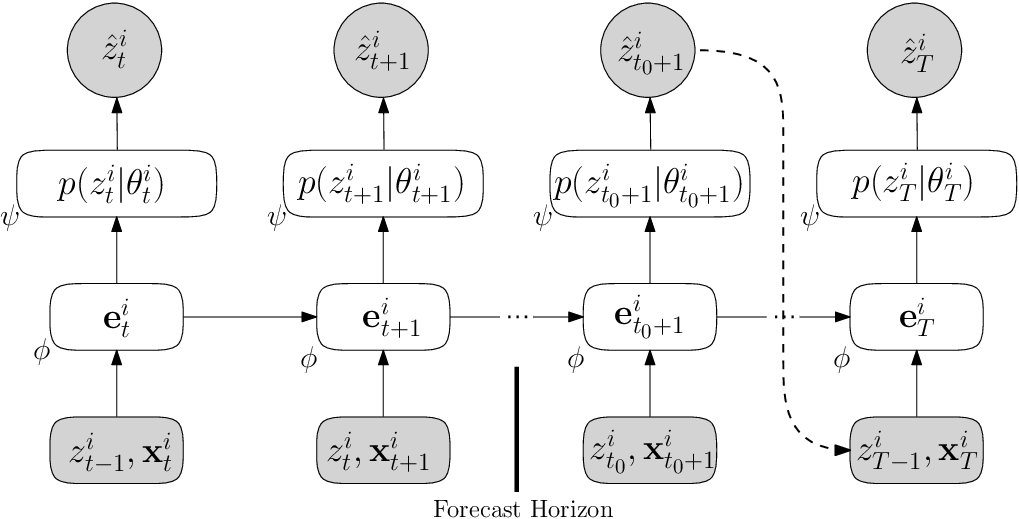}}
			\caption{Unrolled base architecture. On the left of the forecast horizon, the conditioning range can be found $\{1 : t_{0}-1\}$. On its right, the forecasting range $\{t_0 : t_{0}+\tau\}$.}
			\label{fig:deepar}
		\end{center}
	\end{figure}

	During both training and testing, the conditioning range $\{1 : t_{0}-1\}$, which is analogous to the encoder of seq2seq models \cite{sutskever2014sequence}, transfers information to the forecasting range $\{t_0 : t_{0}+\tau\}$, analogous to the decoder. Therefore, this base framework can be interpreted as an encoder-decoder architecture, with the consideration that both encoder and decoder are the same network, as Fig. \ref{fig:deepar} shows. 
	
	For forecasting, directly sampling from the model can be done as $\hat{z}^{i}_{t_0:t_0+\tau} \sim{~}p\left(z_{t_{0}: t_{0}+\tau}^{i} \mid z_{1: t_{0}-1}^{i} , \mathbf{x}^{i}_{1:t_{0}+\tau}, \theta\right)$, when the model consumes the previous time-step prediction $\hat{z}_{t-1}^{i}$ as input, unlike during the conditioning range, where $z_{t-1}^{i}$ is observed. This is illustrated in Fig. \ref{fig:deepar}.
	
	Note that Transformers, unlike RNNs or LSTMs, do not compute the embedding in a sequential manner. Accordingly, when obtaining the embedding through a Transformer model \cite{wu2020deep} and so to use the encoder-decoder architecture previously described, we use the Transformer decoder-only mode, in\-tro\-du\-ced in \cite{liu2018generating}.
	
	\subsection{Characterizing the time series' embedding in the frequency domain}
	\label{timeseries_rp}
	
	Our approach in this paper exploits information in the spectral domain. In this regard, time series' embedding space can be statistically characterized as instances of a random process for which spectral information can be analyzed using the expected autocorrelation and the Power Spectral Density (PSD) \cite{buttkus2012spectral}.
	


	The power spectrum per embedding's dimension can be calculated from an averaged autocorrelation estimated from a finite number $ M\in  \mathbb{Z^+}$ of time series of duration $ T\in  \mathbb{N}$ each:
	

	\begin{equation} \label{eq:Rx_no_embedding}
	\hat{R}_{d}(\kappa)=\frac{1}{M} \sum_{j=1}^{M}  \left(\frac{1}{T} \sum_{t=1}^{T} e^{j}_t(d) \; e^{j}_{t+\kappa} (d) \right) \in \mathbb{R}^{ T},
	\end{equation}
	where $\hat{R}_{d}(\kappa) \in \mathbb{R}^{T}$ is the expected autocorrelation for the $d$-th di\-men\-sion of the $i$-th embedded sequence, $\boldsymbol{e}^{j}(d)$,  and $\kappa$ is the lag between observations.

	The Blackman-Tukey method \cite{blackman1958measurement}, which takes advantage of the Discrete Fourier Transform (DFT) of a windowed autocorrelation, can be used to estimate the PSD once the auto\-co\-rre\-la\-tion has been computed:
	
	\begin{equation} \label{eq:blackman}
	\hat{S}_{d}(\omega) = \sum_{-T}^{T} \hat{R}_{d}(\kappa) e^{-j \omega \kappa} \in \mathbb{R}^{N_{\mathcal{F} \mathcal{T}}},
	\end{equation} 
	where $\hat{S}_d(\omega) \in \mathbb{R}^{N_{\mathcal{F} \mathcal{T}}}$ decomposes a function into its cons\-ti\-tuent frequencies ($N_{\mathcal{F} \mathcal{T}}$ is usually equal to the time series' length $T$), obtaining the estimated spectrum $\hat{S}_d(\omega)$ of the random process that generates the time series in the embedding space. Notice that $\hat{S}_d(\omega)$ is a spectral characterization of the $d$-dimension of the embedding space, hence not global to the process. In Section \ref{FAAM}, we propose an alternative auto\-co\-rre\-la\-tion function that considers each embedding's dimensions as independent realizations of the same random process that generates the time series, therefore obtaining a global spectral representation of the process.
	
	
	

	

   \section{Spectral Attention Autoregressive Model}
	\label{FAAM}
	
	In this section, we introduce the Spectral Attention Autoregressive Model (SAAM), a general framework that incorporates a Spectral Attention (SA) module able to exploit embedding function's spectral information using attention mechanisms to solve two main tasks:
	
	\hfil
	
	\begin{enumerate}
		\item Time series $z_{t}^{i}$ are governed by global trends and sea\-so\-na\-li\-ty structures. SAAM captures these global patterns and incorporates them into the forecast.
		\item Time series exhibit noise around these primary trends that difficult the forecasting process. SAAM filters this noise using spectral filtering, improving the signal to noise ratio thus alleviating the error propagation problem autoregressive models suffer from.
	\end{enumerate}

	\hfil

	These two operations, incorporating global trends into the forecast and filtering the time series' local context, are encapsulated in the SA module, responsible of all the frequency domain related operations. As such, it can be incorporated in any deep-autoregressive structure.
		
\begin{figure}[h] 
		\begin{center}
			\centerline{\includegraphics[width=0.65\columnwidth]{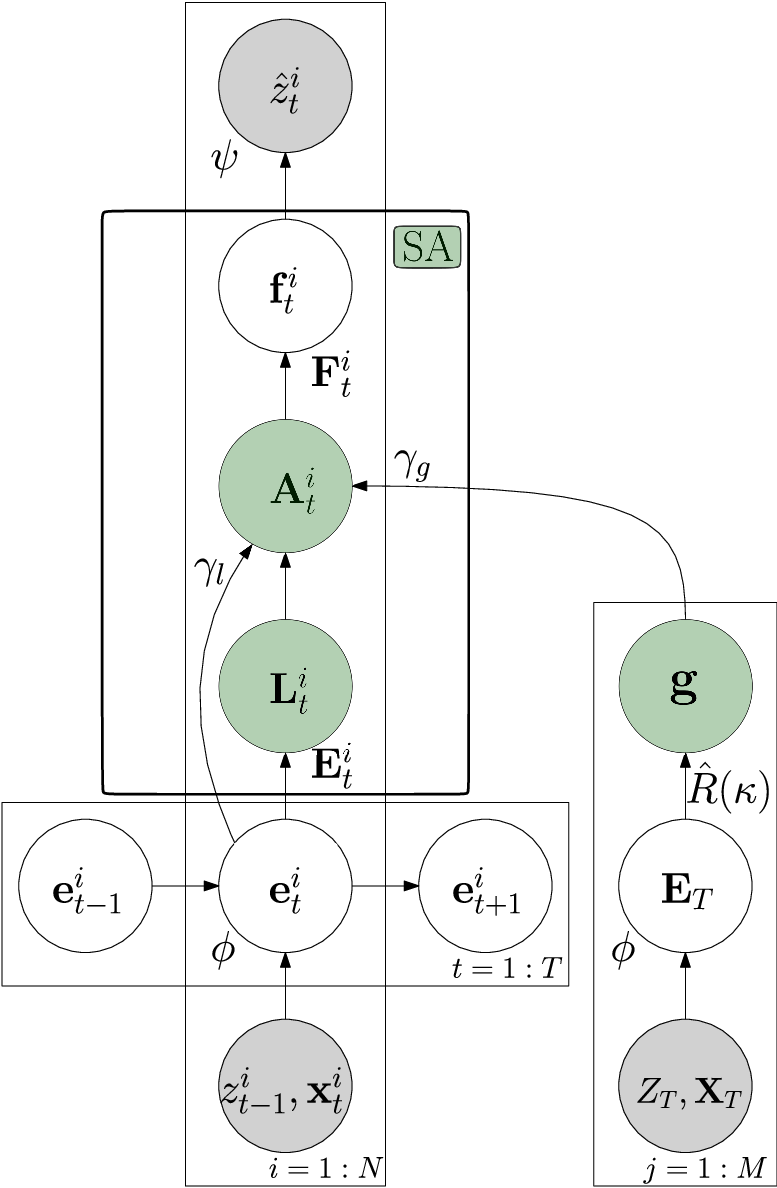}}
			\caption{Spectral Attention Autoregressive Model general architecture. Gray represents observed variables. Green indicates frequency domain re\-pre\-sen\-ta\-tions.}
			\label{fig:general_architecture}
		\end{center}
	\end{figure}

	
	The resulting architecture, displayed in Fig. \ref{fig:general_architecture}, is therefore composed by three main parts: embedding function, SA module and probabilistic model.
	For further details on the embedding and probabilistic model, common to the base architecture of Fig. \ref{fig:basic_architecture}, we refer the reader to Section \ref{base_architecture}. With respect to the SA module, we now explain in detail how it integrates both global and local information into the forecasting.

		\begin{table*}[]
			
				\fontsize{11}{9}\selectfont
			\centering
			\caption{Summary of operations performed by SAAM to filter both local and global contexts..}
			\label{eq:fam}
			\begin{align*}
			\text { Embedding function: }  \mathbf{e}_{t}^{i}&=f_{\phi}\left(\mathbf{e}_{t-1}^{i}, z^{i}_{t-1}, \mathbf{x}_{t}^{i}\right), \in \mathbb{R}^{D}  \\[1pt]
			\text { Buffered embedding: } \mathbf{E}_{t}^{i}&=\mathbf{e}_{t-T_F : t}^{i}, \in \mathbb{R}^{D \times T_F} \\[2pt]
			\text { Local spectrum: }  \mathbf{L}_{t}^{i}&=\vert \mathcal{F} \mathcal{T}\left\{ \mathbf{E}_{t}^{i}\right\}\vert,  \in \mathbb{R}^{D \times N_{\mathcal{F} \mathcal{T}}}\\[2pt]
			\text { Global spectrum: } \mathbf{g}&=\mathcal{F} \mathcal{T}\left\{\hat{R}_{T}(\tau)\right\}, \in \mathbb{R}^{N_{\mathcal{F} \mathcal{T}}} \\[2pt]
			\text { Local filtering: }  \boldsymbol{\alpha}_{l, t}^{i}&={f}_{\gamma_l}({\mathbf{e}}_{t}^{i}, \mathbf{L}_{t}^{i}), \in \mathbb{R}^{D \times N_{\mathcal{F} \mathcal{T}}}  \\[2pt]
			\text { Global filtering: } \boldsymbol{\alpha}_{g, t}^{i}&={f}_{\gamma_g}({\mathbf{e}}_{t}^{i}, \mathbf{G}), \in \mathbb{R}^{D \times N_{\mathcal{F} \mathcal{T}}}  \\[2pt]
			\text { Spectral attention: }  \mathbf{A}_{t}^{i}&=\mathbf{L}_{t}^{i} \odot  \boldsymbol{\alpha}_{l, t}^{i} +\mathbf{G}_{t}^{i} \odot  \boldsymbol{\alpha}_{g, t}^{i}, \in \mathbb{R}^{D \times N_{\mathcal{F} \mathcal{T}}} \\[2pt]
			\text { Filtered embedding: } \mathbf{F}_{t}^{i}&=\mathcal{F} \mathcal{T}^{-1} \left\{  \mathbf{A}_{t}^{i} \right\},  \in \mathbb{R}^{D \times T_F} \\[2pt]
			\text { Time-step filtered embedding: }\mathbf{f}_{t}^{i}&=\mathbf{F}_{t}^{i}(t), \in \mathbb{R}^{D} \\[2pt]
			\text { Emission : } \hat{z}_{t}^{i} & \sim p\left(z_{t}^{i} \mid \mathbf{f}_{t}^{i}, \theta \right), \in \mathbb{R}
			\end{align*}
	\end{table*}

	 \subsection{Global Spectral Information}
	 \label{global}

	 To incorporate global patterns into the prediction, we perform a frequency domain characterization of the process, as explained in Section \ref{timeseries_rp}. We aim to exploit information in the spectral domain associated with the neural embeddings $\left\{\mathbf{e}_{1}^{i}, \mathbf{e}_{2}^{i}, \ldots, \mathbf{e}_{t_0 - 1}^{i}\right\}_{i=1}^{M}$ of a number $M \in  \mathbb{N}$ of time series.
	 
	 Nevertheless, we are not interested in characterizing each embedding's dimension independently, $\hat{S}_d(\omega) \in \mathbb{R}^{N_{\mathcal{F} \mathcal{T}}}$, as using Eq. \ref{eq:Rx_no_embedding} entails. This would require a mul\-ti\-di\-men\-sional PSD for characterizing the process over all embedding's dimensions, $\hat{S}(\omega) \in \mathbb{R}^{D \times N_{\mathcal{F} \mathcal{T}}}$.
	 Instead, we consider each dimension of the em\-be\-dding as independent realizations of the same process we average over, resulting in the following autocorrelation function:

	 	\begin{equation} \label{eq:Rx_complete}
	 \hat{R}(\kappa)=\frac{1}{M} \sum_{j=1}^{M}  \left[ \frac{1}{D} \sum_{d=1}^{D}   \left(\frac{1}{T} \sum_{t=1}^{T} e^{j}_t(d) \; e^{j}_{t+\kappa} (d) \right)\right] \in \mathbb{R}^{T},
	 \end{equation}	
	 where ${e}_{t}^{i}(d)$ is the $d$-th dimension of the $j$-th embedded sequence and $D \in \mathbb{Z^+}$ the embedding's number of dimensions. To apply the Blackman-Tukey method from Eq. \ref{eq:blackman}  over the previous autocorrelation function involves to obtain the complete frequency-domain characterization of the process, in\-cor\-po\-ra\-ting all the global patterns across the different dimensions of the embedding into a single spectral representation, $\hat{S}(\omega) \in \mathbb{R}^{N_{\mathcal{F} \mathcal{T}}}$.


	 Therefore, Eqs. \ref{eq:blackman} and \ref{eq:Rx_complete} allow us to calculate a Monte Carlo approximation of the process' PSD for a batch of $M\in  \mathbb{N}$ trai\-ning sequences $\{Z_{T}, \mathbf{X}_{T} \}_{j=1}^{M}$, via a Fourier Transform with $N_{\mathcal{F} \mathcal{T}}$ points: $ \mathbf{g}=\mathcal{F} \mathcal{T}\left\{\hat{R}_{T}(\tau)\right\} \in \mathbb{R}^{N_{\mathcal{F} \mathcal{T}}}$. 
	 Consequently, $\mathbf{g} \in \mathbb{R}^{N_{\mathcal{F} \mathcal{T}}} $ is the global spectral representation of the process, shared among the $N \in \mathbb{N}$ time series to forecast. 
	
	 \subsection{Local Spectral Information}
	 \label{local}
	 
	 To perform the spectral local filtering, we analyze the last $T_F \in \mathbb{N}$ values of each individual sample: $\mathbf{E}_{t}^i  = \left\{\mathbf{e}_{t-T_F}^{i}, \mathbf{e}_{t-T_F+1}^{i}, \ldots, \mathbf{e}_{t}^{i}\right\} \in \mathbb{R}^{D \times T_F}$ encapsulates the previous  $T_F$ embeddings, each of dimension $ \mathbb{R}^{D}$. This embedded buffered signal at time $t$ is transformed to the frequency domain, $ \mathbf{L}_{t}^{i}=\vert \mathcal{F} \mathcal{T}\left\{ \mathbf{E}_{t}^{i}\right\}\vert  \in \mathbb{R}^{D \times N_{\mathcal{F} \mathcal{T}}}$, via a DFT with $N_{\mathcal{F} \mathcal{T}}$ points. Note that we retain only the module of the Fourier Transform.

	  \subsection{Merging Global and Local Contexts  with Spectral Attention}
	 \label{merging}
	
	We combine both global and local spectral information to modify the embedded representation of the time series $\mathbf{e}_{t}^{i}$.  This is done through two spectral attention models contained in the SA module, with parameters $\gamma=\left\{\gamma_g, \gamma_l\right\}$: 
	
	\hfil
	
	\subsubsection{Global Spectral Attention}
	
	this frequency-domain attention model, with parameters $\gamma_g$, is responsible of in\-cor\-po\-ra\-ting, at time $t\in T$ and for each time series, the Global Spectral Information into the forecast. To do so, it uses  the time series' local context, summarized via the embedding function $\mathbf{e}_{t}^{i}$, as key to select the relevant frequency components on $\mathbf{G}$ that should be included during the prediction, where $\mathbf{G} \in \mathbb{R}^{D \times N_{\mathcal{F} \mathcal{T}}}$ is a repetition of $\mathbf{g} \in \mathbb{R}^{N_{\mathcal{F} \mathcal{T}}}$ along the $D$ dimensions of the embedding: $\boldsymbol{\alpha}_{g, t}^{i}={f}_{\gamma_g}({\mathbf{e}}_{t}^{i}, \mathbf{G}) \in \mathbb{R}^{D \times N_{\mathcal{F} \mathcal{T}}} $. The global filter's coefficients, $\boldsymbol{\alpha}_{g, t}^{i}$, take values $\in [0, 1]$ and ${f}_{\gamma_g}(\cdot)$ is a neural network.
	

	\subsubsection{Local Spectral Attention}
		
	as in the Global Spectral Attention model, the embedded representation $\mathbf{e}_{t}^{i}$ is used as key to determine the relevant and not relevant local spectral components, $\boldsymbol{\alpha}_{l, t}^{i}={f}_{\gamma_l}({\mathbf{e}}_{t}^{i}, \mathbf{L}_{t}^{i}) \in \mathbb{R}^{D \times N_{\mathcal{F} \mathcal{T}}}$, where $\gamma_l$ are the Local Spectral Attention's parameters and $\boldsymbol{\alpha}_{l, t}^{i} \in [0, 1]$ are the local filter's coefficients for each embedding's dimension.
	
	\hfil

	We combine both spectral-domain attention models through multiplication and addition operations in the frequency domain: $\mathbf{A}_{t}^{i}=\mathbf{L}_{t}^{i} \odot  \boldsymbol{\alpha}_{l, t}^{i} + \mathbf{G}_{t}^{i} \odot  \boldsymbol{\alpha}_{g, t}^{i} \in \mathbb{R}^{D\times N_{\mathcal{T} \mathcal{F}}} $. 
	The multiplication over the embedding's local spectrum representation $\mathbf{L}_{t}^{i} \odot  \boldsymbol{\alpha}_{l, t}^{i}$ is performing the local spectral filtering, setting to zero not re\-le\-vant frequency components through the local attention model. Furthermore, the addition of $\mathbf{G}_{t}^{i} \odot  \boldsymbol{\alpha}_{g, t}^{i}$ includes significant global trends into the forecast via $\boldsymbol{\alpha}_{g, t}^{i}$, which selects relevant patterns from the process' global spectrum representation $\mathbf{G}_{t}^{i}$.

	Finally, this spectral representation is transformed back to the time domain $\mathbf{F}_{t}^{i}=\mathcal{F} \mathcal{T}^{-1} \left\{  \mathbf{A}_{t}^{i} \right\}  \in \mathbb{R}^{D \times T_F}$ and the last value of $\mathbf{F}_{t}^{i}$, $\mathbf{f}_{t}^{i} \in \mathbb{R}^{D}$, is feed to the probabilistic model to forecast the next time-step, $\hat{z}_{t}^{i} \sim p\left(z_{t}^{i} \mid \mathbf{f}_{t}^{i}, \theta \right)$.

\begin{figure}[ht] 
	 	\begin{center}
	 		\centerline{\includegraphics[width=1.0\columnwidth]{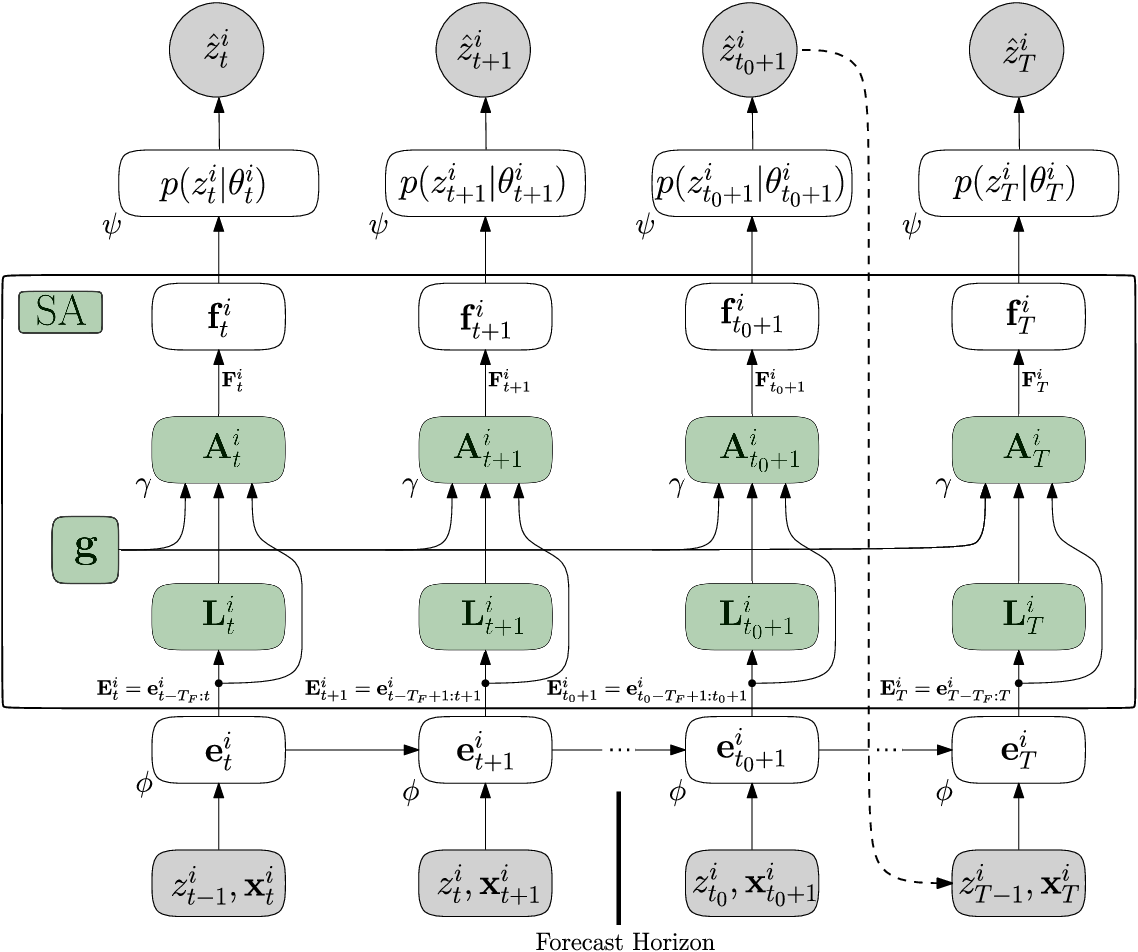}}
	 		\caption{The unrolled architecture of the Spectral Attention Autoregressive Model.}
	 		\label{fig:deepar_fam}
	 	\end{center}
	 	\vskip -0.2in
	 \end{figure}

	 Fig. \ref{fig:deepar_fam} shows the unrolled architecture of the proposed model. Removing the SA module, the base framework displayed in Fig. \ref{fig:deepar} remains.
	 Note that both the computation of the expected autocorrelation and the Fourier Transform  are differentiable with respect to the model parameters.
	 
	\subsection{Training}
	\label{training}


 	The likelihood of the model, which is maximized to match the statistical properties of the data, now includes SA's parameters, $\gamma$:
 	
 	
 	\begin{equation}
 	\begin{aligned}
 	\mathcal{L}(\theta)=\sum_{i=1}^{N} \sum_{t=t_0
 	 }^{t_0-1} p\left(z_{t}^{i} \mid \mathbf{x}_{1:t-1}^{i},\theta (\phi, \gamma, \psi) \right).
 	\end{aligned}
 	\end{equation}

 	Once parameters $\theta$ have been learned during the conditioning range, forecast can be produced as in Section \ref{base_architecture}: $\hat{z}^{i}_{t_0:t_0+\tau} \sim{~}p\left(z_{t_{0}: t_{0}+\tau}^{i} \mid z_{1: t_{0}-1}^{i} , \mathbf{x}^{i}_{1:t_{0}+\tau}, \theta\right)$, computing the joint distribution over the forecasting range for each time series.

 	Notice that, during both training and testing, global spectral information is obtained using a mini-batch of time series different to the one being forecasted.

	\section{Experiments}
	\label{experiments}
	
	We conduct experiments with both synthetic and real-world datasets in order to provide evidence of the superior forecasting ability of SAAM. Moreover, ablation studies are conducted. The code for our proposed model is publicly available on GitHub \footnote{\url{https://github.com/fmorenopino/SAAM}} \cite{fernando_moreno_pino_2021_5086179}.
	
	\subsection{Applying the Spectral Attention Module to State-of-the-Art Models}
	\label{applying}
	
	Many well-known deep-autoregressive models that define the state-of-the-art can take advantage of the Spectral Attention module proposed in Section \ref{FAAM}. This allows them to filter not relevant spectral components and to recognize global trends that can be incorporated into the forecast. 
	
	We integrated the SA module into two forecasting models: DeepAR \cite{salinas2020deepar}, a widely-known and largely used model with se\-ve\-ral industrial applications \cite{schelter2018automating,bose2017probabilistic,schelter2018challenges,fildes2019retail,alexandrov2019gluonts}, which employs a LSTM to perform the embedding; and ConvTrans  \cite{li2019enhancing}, a more recent Transformer-based proposal, which constitutes one of most efficient approaches for using Transformers for the problem of time series forecasting.
	
	We would like to remark that the added complexity by the Spectral Attention module is equivalent to classic attention models as  \cite{bahdanau2014neural}. In these models, for a sequence of length $L$, computing scores between every pair causes $O(L^2)$ memory use. The complexity of Spectral Attention depends on the embedding's dimension $D \in \mathbb{Z^+}$ and the number of points used to compute the Fourier Transform $N_{\mathcal{F} \mathcal{T}} \in \mathbb{Z^+}$. The complexity therefore, depends on $O(D \cdot N_{\mathcal{F} \mathcal{T}})$.

	\subsection{Metrics}
	\label{metrics}
	
	The normalized $\rho-$quantile loss, $QL_{\rho}$, with $\rho \in (0, 1)$, which quantifies the accuracy of a quantile $\rho$ of the predictive distribution, is the main metric used to report the results of the experiments, as in many other works \cite{salinas2020deepar}, \cite{li2019enhancing},  \cite{rangapuram2018deep}, \cite{seeger2016bayesian}.
	
		\begin{equation}
	\begin{split}
	\mathrm{QL}_{\rho}(\boldsymbol{z}, \hat{\boldsymbol{z}})&=2 \frac{\sum_{i, t} P_{\rho}\left(z_{t}^{(i)}, \hat{z}_{t}^{(i)}\right)}{\sum_{i, t}\left|z_{t}^{(i)}\right|},\\
	\quad P_{\rho}(z, \hat{z})&=\left\{\begin{array}{ll}
	\rho(z-\hat{z}) & \text { if } z>\hat{z}, \\
	(1-\rho)(\hat{z}-z) & \text { otherwise }.
	\end{array}\right.
	\end{split}
	\end{equation}
	
	Rolling window predictions after the last point seen in the conditioning range are used to obtain the results. To compute this metric, we use 200 samples from the decoder to estimate $\hat{z}^{i}_{t}$ along time. Also, the normalized sum of the quantile losses is considered, as can be appreciated in the previous equation. 
	
	The Normalized Deviation (ND) \cite{yu2016temporal} and Root Mean Square Error (RMSE) were also used to evaluate the pro\-ba\-bi\-lis\-tic forecasts, especially on the synthetic dataset experiments:
	
	\begin{equation}
	\begin{split}
	\mathrm{ND} &=\frac{\sum_{i, t}\left|z_{i, t}-\hat{z}_{i, t}\right|}{\sum_{i, t}\left|z_{i, t}\right|}, \\[5pt]
	\text { RMSE } &=\frac{\sqrt{\frac{1}{N\left(T-t_{0}\right)} \sum_{i, t}\left(z_{i, t}-\hat{z}_{i, t}\right)^{2}}}{\frac{1}{N\left(T-t_{0}\right)} \sum_{i, t}\left|z_{i, t}\right|}.\\[5pt]
	\end{split}
	\end{equation}

	\subsection{A Synthetic Dataset}
	\label{synthetic}
	
\begin{figure}[h] 
	\begin{center}
		\centerline{\includegraphics[width=0.9\columnwidth]{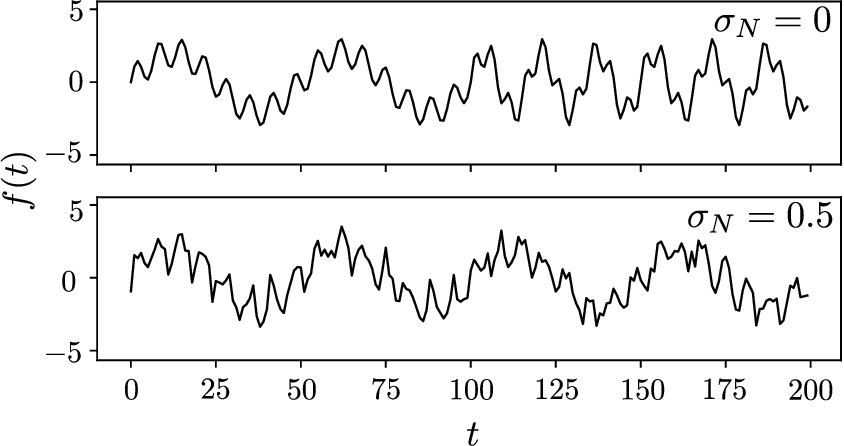}}
		\caption{Synthetic dataset example signal, generated with $f_1=2$, $f_2=15$, $f_3=6$, $f_4=20 \:Hz$. For each signal, each half randomly varies according to the Bernoulli distribution. Noise's standard deviation is increased from top to bottom.} 
		\label{fig:examples_synthetic}
	\end{center}
\end{figure}

	In order to demonstrate SAAM capabilities, we conducted experiments on synthetic data composed of sinusoidal signals with no covariates and a duration of 200 samples. Each of these signals is divided into two halves, randomly selecting the components for each of them as:
	
	\comment{
		\begin{equation}
		\label{eq:synthethic}
		f(t)=\left\{\begin{array}{ll}
			A_{1} \sin (2 \pi f_1 t) + A_{2} \sin (2 \pi f_2 t) + N,  & t \in[0, 100) \\
			A_{3} \sin (2 \pi f_3 t) + A_{4} \sin (2 \pi f_4 t) + N,  & t \in[100, 200],
		\end{array}\right.
		\end{equation}
	}
	
		\begin{equation}
	\label{eq:synthethic}
	f(t)=\left\{\begin{array}{ll}
			\left\{\begin{array}{ll}
				f \sim f_1(t), \text{ if } x_1=0\\
				f \sim f_2(t), \text{ if } x_1=1
			\end{array}\right., & t \in[0, 100), \\
			\\
			\left\{\begin{array}{ll}
				f \sim f_1(t), \text{ if } x_2=0\\
				f \sim f_2(t), \text{ if } x_2=1
			\end{array}\right., & t \in[100, 200].
	\end{array}\right.
	\end{equation}
	 where $x_1$ and $x_2$ are independently and randomly generated from a Bernoulli distribution, $X \sim Ber(\theta)$ with probability $\theta = 1/2$. This implies that each half of the time series can take the form of one over two signals, $f_1(t)$ or $f_2(t)$, both of them composed by the addition of two sines of different frequencies plus a noise component, $\nu$:
	
	\begin{equation}
		\label{eq:synthethic}
		\left\{\begin{array}{ll}
		f_1(t)=A_{1} \sin (2 \pi f_1 t) + A_{2} \sin (2 \pi f_2 t) + \nu, \\
		f_2(t)=A_{3} \sin (2 \pi f_3 t) + A_{4} \sin (2 \pi f_4 t) + \nu,
		\end{array}\right.
	\end{equation}
	where $\nu \sim \mathcal{N}(0, \sigma^{2}_{\nu})$ and $\sigma^{2}_{\nu}$ will vary during the experiments, the amplitudes had fixed values, $A_1, A_2, A_3, A_4 = 2$ and each of the frequencies used are chosen from a different interval as: $f_1 \sim [1,5], f_2 \sim [15,20], f_3 \sim[5, 10], f_4 \sim [20, 25]$. A sampled time series is shown in Fig. \ref{fig:examples_synthetic}.

	To evaluate spectral attention advantages we trained two models on this dataset: 1) DeepAR as base model and 2) SAAM. Both of them used the exact same architecture for the common parts: the embedding function was performed by a LSTM of 3 layers and 10 hidden units per layer and the probabilistic model is composed of a fully connected network.


	\hfil
	
	\subsubsection{Empirical Analysis}
	
\begin{figure}[h] 
		\begin{center}
			\centerline{\includegraphics[width=1.0\columnwidth]{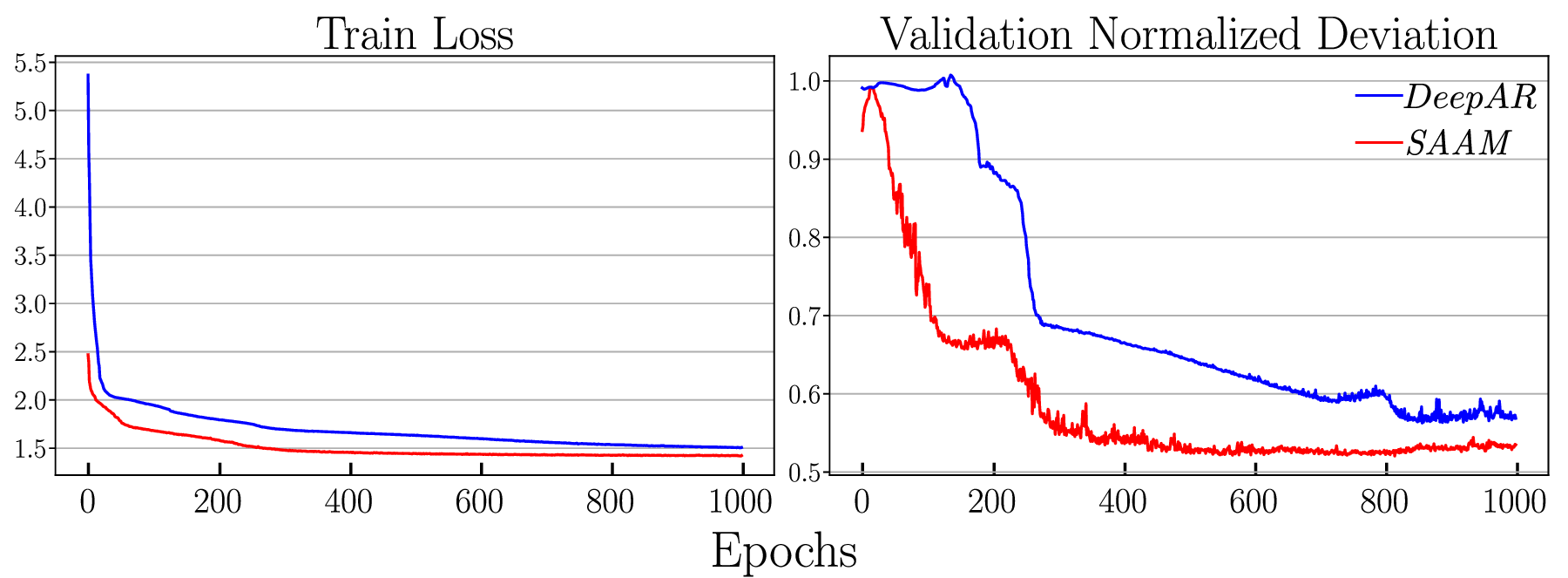}}
			\caption{Training loss (left) and validation ND (right). The later for a forecast window $\tau=50$.} 
			\label{fig:train_loss_nd}
		\end{center}
	\vskip -0.15in
	\end{figure}

	 Both models DeepAR and SAAM were trained using 500 signals from the synthetic dataset with a noise com\-po\-nent $N \sim \mathcal{N}(0, \sigma^{2}_{\nu}=0.5)$. The training loss evolution and the validation ND error are shown in Fig. \ref{fig:train_loss_nd}
	
	We then evaluated the trained models in two different scenarios. First, in absence of a noise component, $N \sim \mathcal{N}(0, \sigma^{2}_{\nu}=0)$; second, in the same conditions they were trained, $N \sim \mathcal{N}(0, \sigma^{2}_{\nu}=0.5)$.
	
	Moreover, for each of the previous cases, we evaluated the models using different forecast horizons, starting with $\{t_0=175, \tau=25\}$ up to  $\{t_0=105, \tau=95\}$. In this last case, the final 95 time-steps are forecasted after observing the five first samples from the time series' second half. Setting $t_0 \le 100$ would make no sense, as $t \in [0, 100)$ contains no evidence about the time series configuration during $t \in [100, 200]$.

\begin{table*}[]
	\fontsize{6.8}{10}\selectfont
	\centering
	\caption{Comparative of DeepAR vs SAAM on the synthetic dataset for different noise component's variance and forecast lengths.}
	\label{table:synthetic}
	\begin{tabular}{ccllcccc}
		\thickhline
		\multicolumn{8}{c}{\textbf{DeepAR / SAAM}}                                                                                                                         \\ \thickhline
		\textbf{$\sigma^{2}_{\nu}$}                &  \textbf{$t_0$} & \multicolumn{1}{c}{\textbf{$\tau$}} &  & \textbf{ND}                                    & \textbf{RMSE}                                  & \textbf{$\mathbf{QL_{\rho=0.5}}$}                            & \textbf{$\mathbf{QL_{\rho=0.9}}$}                                      \\ \cline{1-3} \cline{5-8} 
		\multirow{5}{*}{\textbf{0}}   & \textbf{175} & \textbf{25}                      &  & 0.25494 $\pm$ 0.000 / \textbf{0.24101 $\pm$ 0.003} & 0.35004 $\pm$ 0.000 / \textbf{0.28680 $\pm$ 0.003} & 0.26250 $\pm$ 0.000 / \textbf{0.24872 $\pm$ 0.004} & \textbf{0.21830 $\pm$ 0.000} / 0.221701 $\pm$ 0.003 \\
		& \textbf{150} & \textbf{50}                      &  & 0.27403 $\pm$ 0.000 / \textbf{0.20145 $\pm$ 0.002}  & 0.37677 $\pm$ 0.000 / \textbf{0.24868 $\pm$ 0.004}  & 0.27882 $\pm$ 0.000 / \textbf{0.20594 $\pm$ 0.002}  & 0.26836 $\pm$ 0.000 / \textbf{0.26325 $\pm$ 0.004}           \\
		& \textbf{120} & \textbf{80}                      &  & 0.37867 $\pm$ 0.000 / \textbf{0.17323 $\pm$ 0.004}  & 0.47434 $\pm$ 0.000 / \textbf{0.21293 $\pm$ 0.006}  & 0.38511 $\pm$ 0.000 / \textbf{0.17811 $\pm$ 0.005}  & 0.38712 $\pm$ 0.000 / \textbf{0.13148 $\pm$ 0.001}           \\
		& \textbf{110} & \textbf{90}                      &  & 0.75234 $\pm$ 0.000 / \textbf{0.20326 $\pm$ 0.001} & 1.06398 $\pm$ 0.000 / \textbf{0.28207 $\pm$ 0.001} & 0.75810 $\pm$ 0.001 / \textbf{0.20828 $\pm$ 0.001} & 0.73909 $\pm$ 0.001 / \textbf{0.14980 $\pm$ 0.003}           \\
		& \textbf{105} & \textbf{95}                      &  & 0.82233 $\pm$ 0.000 / \textbf{0.37575 $\pm$ 0.003} & 1.13000 $\pm$ 0.000 / \textbf{0.66817 $\pm$ 0.001} & 0.82801 $\pm$ 0.001 / \textbf{0.38062 $\pm$ 0.004} & 0.80492 $\pm$ 0.001 / \textbf{0.34665 $\pm$ 0.002}           \\ \cline{1-3} \cline{5-8} 
		\multirow{5}{*}{\textbf{0.5}} & \textbf{175} & \textbf{25}                      &  & 0.58778 $\pm$ 0.000 /  \textbf{0.56469 $\pm$ 0.002} & 0.75065 $\pm$ 0.000 / \textbf{0.71553 $\pm$ 0.002} & 0.59492 $\pm$ 0.001 / \textbf{0.57267 $\pm$ 0.002} & 0.59893 $\pm$ 0.001 / \textbf{0.61914 $\pm$ 0.005}           \\
		& \textbf{150} & \textbf{50}                      &  & 0.60786 $\pm$ 0.000 / \textbf{0.55649 $\pm$ 0.001} & 0.76756 $\pm$ 0.000 / \textbf{0.70871 $\pm$ 0.002} & 0.61322 $\pm$ 0.000 / \textbf{0.56057 $\pm$ 0.001} & 0.61094 $\pm$ 0.000 / \textbf{0.60490 $\pm$ 0.003}           \\
		& \textbf{120} & \textbf{80}                      &  & 0.63071 $\pm$ 0.000 / \textbf{0.57218 $\pm$ 0.002} & 0.79379 $\pm$ 0.000 / \textbf{0.73353 $\pm$ 0.002} & 0.63531 $\pm$ 0.001 / \textbf{0.57674 $\pm$ 0.002} & 0.62907 $\pm$ 0.001 / \textbf{0.52398 $\pm$ 0.002}           \\
		& \textbf{110} & \textbf{90}                      &  & 0.78535 $\pm$ 0.000 / \textbf{0.67392 $\pm$ 0.001} & 1.03135 $\pm$ 0.000 / \textbf{0.89065 $\pm$ 0.001} & 0.79090 $\pm$ 0.001 / \textbf{0.67823 $\pm$ 0.001} & 0.79474 $\pm$ 0.001 / \textbf{0.61661 $\pm$ 0.001}           \\
		& \textbf{105} & \textbf{95}                      &  & 0.86135 $\pm$ 0.000 / \textbf{0.70326 $\pm$ 0.009} & 1.12157 $\pm$ 0.000 / \textbf{0.93480 $\pm$ 0.012} & 0.86594 $\pm$ 0.001 / \textbf{0.70809 $\pm$ 0.009} & 0.85522 $\pm$ 0.001 / \textbf{0.65422 $\pm$ 0.012}  
   \\ \thickhline
	\end{tabular}
\end{table*}

\begin{table}[]
	\fontsize{8.0}{12}\selectfont
	\centering
	\caption{DeepAR \& SAAM accuracy deprecation com\-pa\-ra\-tive between $t_0=175/\tau=25$ and $t_0=105/\tau=95$. }
	\label{table:synthetic_comparative}
	\begin{tabular}{cccc }
		\thickhline
		&               & $\mathbf{\sigma^{2}_{\nu}=0}$ & $\mathbf{\sigma^{2}_{\nu}=0.5}$  \\ \thickhline
		& \textbf{ND}   & -222,56 \%    & -46,55 \%               \\
		& \textbf{RMSE} & -222,82 \%     & -49,41 \%                \\
		& \textbf{$\mathbf{QL_{\rho=0.5}}$}  & -215,43 \%     & -45,56 \%              \\
		\multirow{-4}{*}{\textbf{DeepAR}} & \textbf{$\mathbf{QL_{\rho=0.9}}$}  & -268,72 \%      & -42,79 \%               \\ \hline
		& \textbf{ND}   & -55,91 \%     & -24,54 \%              \\
		& \textbf{RMSE} & -132,97 \%     & -30,64 \%              \\
		& \textbf{$\mathbf{QL_{\rho=0.5}}$}  & -53,03 \%             & -38,34\%            \\
		\multirow{-4}{*}{\textbf{SAAM}}   & \textbf{$\mathbf{QL_{\rho=0.9}}$} & -56,36 \%      & -5,67 \%                \\ \thickhline
	\end{tabular}
\end{table}

	\hfil
	
	Table \ref{table:synthetic} shows the results reported by DeepAR and SAAM. After training the models, 10 evaluations per sce\-na\-rio were conducted during testing. On average, SAAM improved the $ND$ by a 28.4\%, $RMSE $ by a 27.7\%, $QL_{\rho=0.5}$ by a 28.2\% and $QL_{\rho=0.9}$ by a 28.4\%, proving that Spectral Attention inclusion enhances the model's performance. 
	
	Furthermore, remark that the Global Spectral Attention model contained in the SA module helps accelerate global trends detection: the earlier the forecast window starts (smaller $t_0$), the better the results of SAAM with respect to the DeepAR base model are.
	Specifically, for $N \sim \mathcal{N}(0, 0)$ and $t_0=175$ (the most favorable setting as there is no noise and just 25 data points to forecast), SAAM and DeepAR report very similar results while, with an increased forecast window of $t_0=105$, SAAM architecture thoroughly enhances base model's results: $ND$ is improved by a 54.3\%, $RMSE $ by a 40.9\%, $QL_{\rho=0.5}$ by a 54\% and $QL_{\rho=0.9}$ by a 56.9\%.
	
	SAAM's ability to detect and incorporate relevant trends into the forecast has direct implications in the reported results. Table \ref{table:synthetic_comparative} shows a comparison for different noise levels of the reported metrics' degradation while increasing the forecast window from $\{t_0 = 175, \tau = 25\}$ to $\{t_0 = 105, \tau = 95\}$. SAAM's deterioration is much smaller than DeepAR's.
	
	Finally, as can be seen on Fig. \ref{fig:train_loss_nd}, a faster training convergence is achieved while using the SA module. Furthermore, SAAM reached the minimum validation error long before DeepAR.

	
	\hfil
	
	\subsubsection{Interpretability Analysis}

\begin{figure}[] 
		\vskip -0.15in
		\centering
		\begin{center}
			\centerline{\includegraphics[width=1.0\columnwidth]{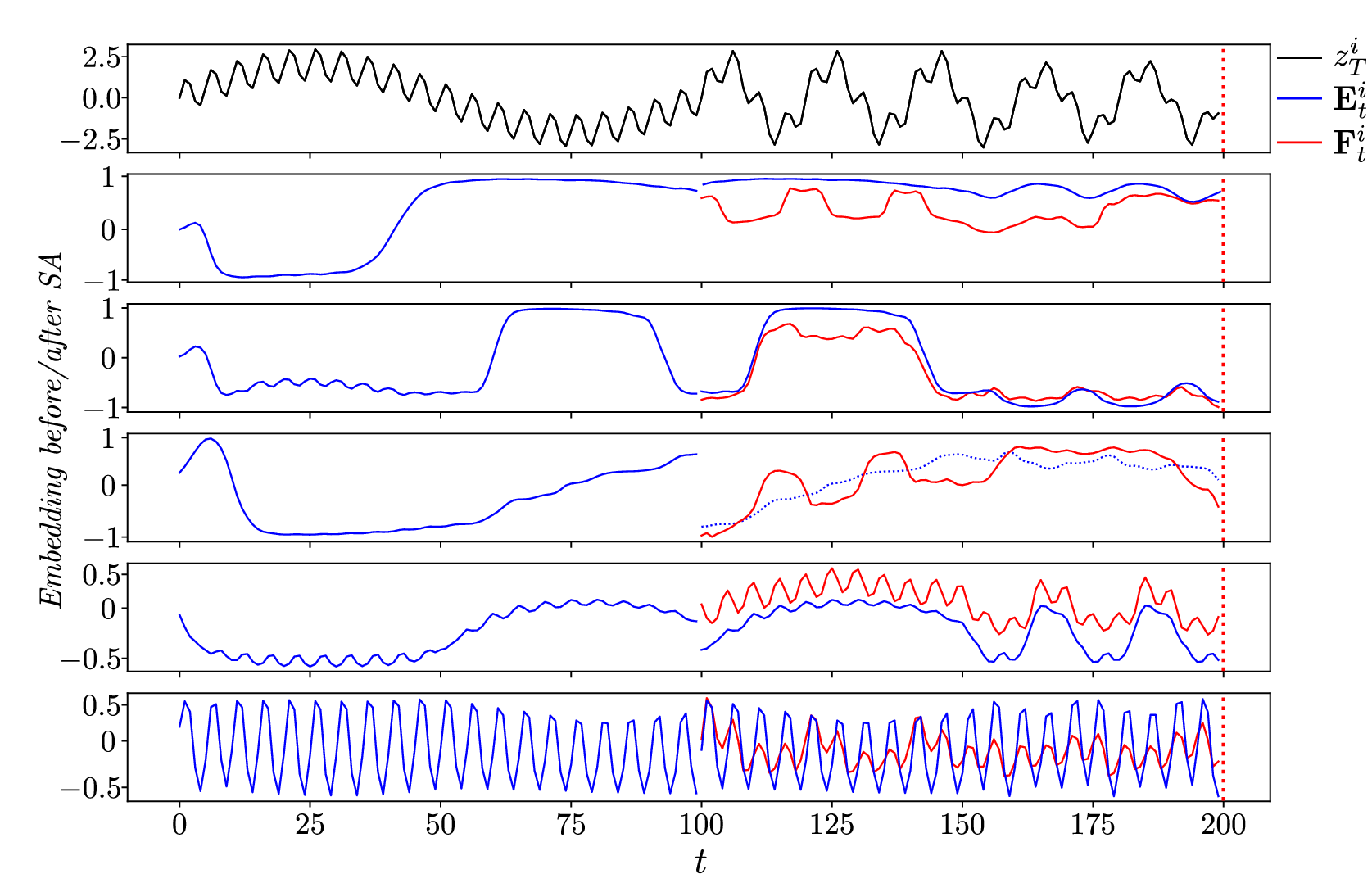}}
			\caption{Hidden representations of SAAM model while forecasting at time $t=200$ (marked by red dotted lines). A filtering window of size $T_F=100$ was used.} 
			\label{fig:hidden_synthetic}
		\end{center}
		
	\end{figure}

	We now visualize how spectral attention affects the embedding evolution during the fo\-re\-cas\-ting, studying SAAM's internal behavior while using a LSTM with 1 layer and 5 hidden units per layer as embedding. In Fig. \ref{fig:hidden_synthetic}, we show the hidden re\-pre\-sen\-ta\-tions produced by SAAM during the forecasting of a time series $z_{T}^{i}$. Each row in this figure represents one of those 5 hidden dimensions at time $t=200$ (predicting time series' final time-step). SAAM's hidden variables are displayed as:
	
	\begin{itemize}
		\item Blue lines represent the hidden representation of the LSTM before SA. This is, the representation that DeepAR would use to forecast.

		\item Red lines represent SA module's output, $\mathbf{F}_{t=200}^{i}$ with dimension $ \left( D \times T_F \right)$, being $D=5$ and $T_F=100$. This is the representation that DeepAR based SAAM uses in order to perform the forecast.
		
		\item We also represent the true signal $z_{T}^{i}$ in the first row of Fig. \ref{fig:hidden_synthetic}, the noise component was removed for clarity. This specific sequence $z_{T}^{i}$ can be described as:

	\end{itemize}
		\begin{equation}
		\label{eq:synthethic_interpretability}
		f(t)=\left\{\begin{array}{ll}
		2\sin (2 \pi t) + 2 \sin (40 \pi t) + \nu, \; t \in[0, 100) \\
		2\sin (10 \pi t) + 2\sin (40 \pi t) + \nu, \; t \in[100, 200] 
		\end{array}\right.
		\end{equation}

	\vskip 0.05in
	
	Fig. \ref{fig:hidden_synthetic} shows how the SA module incorporates global trends into the hidden representation, making the model immediately aware of trend changes: in Dim. 0 and 2, $\mathbf{F}_{T_F}^{i}$ exhibits a trend associated with  $f_3=5Hz$ from time $t \approx 100$, while this component does not appear in $\mathbf{E}_{T}^{i}$ until $t \approx 160$. Furthermore, $\mathbf{F}_{T_F}^{i}$ incorporates in both Dim. 3 and 4 a $20Hz$ component that $z_{T}^{i}$ exhibits for both $f_1(t)$ and $f_2(t)$. 

	These examples show the ability of the proposed architecture to incorporate into the forecast patterns that the time series exhibit.
	
\begin{figure}[] 
		\centering
		\begin{center}
			\centerline{\includegraphics[width=1.0\columnwidth]{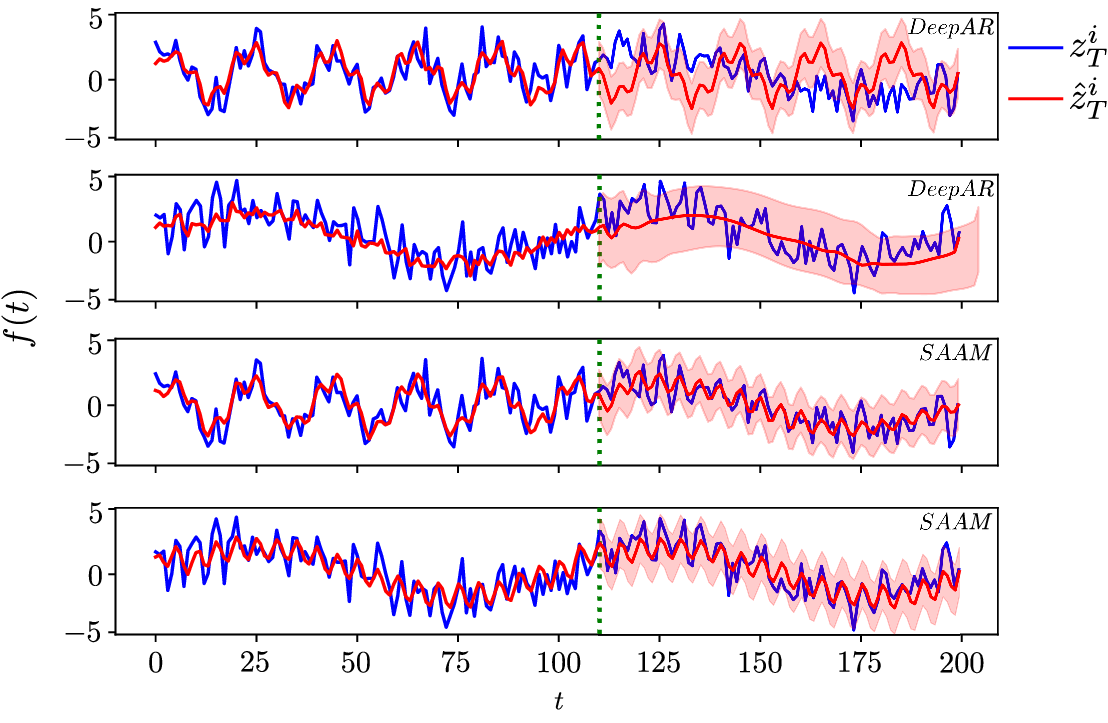}}
			\caption{Synthetic dataset's forecasts examples by DeepAR (top rows) and SAAM (bottom rows) on the same two time series, $N \sim \mathcal{N}(0, \sigma^{2}_{\nu}=0.5)$, $\{t_0 = 110, \tau = 90\}$. Green dotted vertical lines mark the forecast start. Predictions mean and variation appear on red, ground-truth in blue.}
			\label{fig:synthetic_preds}
		\end{center}
	\end{figure}

	Finally, Fig. \ref{fig:synthetic_preds} displays some forecast examples by DeepAR and SAAM. DeepAR frequently failed to detect trend changes at $t=100$ and to incorporate certain frequency variations into the mean and standard deviation of the forecast, while SAAM correctly manages these situations, as Fig. \ref{fig:synthetic_preds} shows. 
	
	\subsection{Real world datasets}
	\label{real_world}

	The performance of SAAM on several real-world datasets was compared with other state-of-the-art models: two classic forecasting techniques, ARIMA \cite{box2008time} and ETS \cite{hyndman2018forecasting}; a recent matrix factorization method, TRMF \cite{yu2016temporal}; a RNN based State Space Model, DSSM  \cite{rangapuram2018deep}; DeepAR \cite{salinas2020deepar} and ConvTrans \cite{li2019enhancing}.
	
	Two different configurations were proposed for SAAM: the first one uses DeepAR as base model, the second combines ConvTrans with SA, both complied with the architecture of Fig. \ref{fig:general_architecture}.
	A basic framework  for the common parts was maintained during all the experiments: for the DeepAR base model, the embedding consisted on 3 LSTM layers with 40 hidden units per layer while, for the ConvTrans base proposal, a Transformer with 8 heads and 3 layers was used. Both set of parameters appear in the original articles \cite{salinas2020deepar},  \cite{li2019enhancing} as optimal choices.
	
	The electricity\footnote{\url{https://archive.ics.uci.edu/ml/datasets/ElectricityLoadDiagrams20112014#}} and traffic datasets \footnote{\url{https://archive.ics.uci.edu/ml/datasets/PEMS-SF}} \cite{Dua:2019}, \cite{NIPS2016_85422afb} were e\-va\-lua\-ted using two different forecast windows, of one and seven days. The electricity dataset contains hourly time series of energy consumption of 370 customers. Similarly, the traffic dataset contains the hourly occupancy rates, with values between zero and one, of 963 car lanes in San Francisco area freeways.

	 Three more datasets, with different forecast windows each, were also used. The solar dataset \footnote{\url{https://www.nrel.gov/grid/solar-power-data.html}} \cite{solar_dataset} contains the solar power production records from January to August 2006 from 137 plants in Alabama and exhibits hourly measurements. Forecast windows of 1 day were predicted during the e\-va\-lua\-tion. The wind dataset \footnote{\url{https://www.kaggle.com/sohier/30-years-of-european-wind-generation}} \cite{wind_dataset} contains daily estimates of 28 countries' wind energy potential in a period from 1986 to 2015, expressed as a percent of a power plant's maximum output. Finally, the M4-Hourly dataset \footnote{\url{https://www.kaggle.com/yogesh94/m4-forecasting-competition-dataset}} \cite{makridakis2018m4}, contains 414 hourly time series from the M4 competition \cite{makridakis2020m4}. Table \ref{table:datasets_details} summarizes each dataset and the models' architecture used.

	 All datasets used covariates $\{\mathbf{x}^{i}_{t}\}_{i=1}^{N} \in \mathbb{R}^{C}$, composed by the hour of the day, day of the week,  week of the year and month of the year, for daily, weekly, monthly and yearly data, respectively. Also, covariates that measure the distance to the first observation of the time series as well as an item index identification for each time series were used.

\begin{table*}[]
	\fontsize{7.5}{11.5}\selectfont
	\centering
	\caption{Datasets evaluated's details.}
	\label{table:datasets_details}
	\begin{tabular}{cccccc}
		\thickhline
		\textbf{}                     & \textbf{Electricity} & \textbf{Traffic} & \textbf{Solar} & \textbf{Wind} & \textbf{M4} \\ \thickhline
		\textbf{Length}               & 32304                & 129120           & 5832           & 10957         & 748         \\ 
		\textbf{\# Time Series}        & 370                  & 370              & 137            & 28            & 414         \\ 
		\textbf{Granularity}          & Hourly               & Hourly           & Hourly         & Daily         & Hourly      \\ 
		\textbf{Domain}               &  $\mathbb{R}$                    & {[}0,1{]}        & $\mathbb{R}$               & $\mathbb{R}$              & $\mathbb{R}$            \\ 
		\textbf{Batch Size}           & 128                  & 128              & 128            & 64            & 128         \\ 
		\textbf{Learning Rate}        & 1e-3                 & 1e-3             & 1e-3           & 1e-3          & 1e-3        \\ 
		\textbf{\# LSTM Layers}        & 3                    & 3                & 3              & 3             & 3           \\ 
		\textbf{\# Hidden Units/Layer} & 40                   & 40               & 40             & 40            & 40          \\ 
		\textbf{\# Heads}              & 8                    & 8                & 8              & 8             & 8           \\ 
		\textbf{\# Layers}            & 3                    & 3                & 3              & 3             & 3           \\ 
		\textbf{Forecast Window}      & 1 Day / 7 Days         & 1 Day / 7 Days     & 1 Day          & 30 Days       & 2 Days      \\ 
		\textbf{Encoder Length}       & 168 / 24                  & 168 / 24              & 168            & 162           & 128         \\ 
		\textbf{Decoder Length}       & 24 / 168               & 24 / 168           & 24             & 30            & 48          \\ \thickhline
	\end{tabular}
\end{table*}

\begin{table*}[]
	\fontsize{7.3}{12}\selectfont
	\centering
	\caption{Evaluations summary, using $QL_{\rho=0.5}$/$QL_{\rho=0.9}$ metrics, on Electricity and Traffic datasets with forecast windows of 1 and 7 days, where \textsuperscript{$\triangle$} are extracted from \cite{li2019enhancing}.}
	\label{table:real_datasets}
	\begin{tabular}{ccccccclcc}
		\thickhline
		\textbf{Dataset}    & \multicolumn{9}{c}{\textbf{Method}}                                                                                                                                            \\ \thickhline
		\textbf{}           & \textbf{ARIMA\textsuperscript{$\triangle$}} & \textbf{ETS\textsuperscript{$\triangle$}} & \textbf{TRMF\textsuperscript{$\triangle$}} & \textbf{DSSM\textsuperscript{$\triangle$}} & \textbf{DeepAR} & \textbf{ConvTrans} &  & \textbf{SAAM (DeepAR)} & \textbf{SAAM (ConvTrans)} \\ \cline{2-7} \cline{9-10} 
		\textbf{elect-1d}   & 0.154/ 0.102    & 0.101 / 0.077  & 0.084 / -       & 0.083 / 0.056   & 0.075 / 0.040     & 0.059 / 0.034        &  & 0.0635 / 0.0317                   & \textbf{0.056} / \textbf{0.029}                        \\
		\textbf{elect-7d}   & 0.283 / 0.109    & 0.121 / 0.101  & 0.087 / -       & 0.085 / 0.052   & 0.082 / 0.053     & 0.079 / 0.051       &  & 0.076 / \textbf{0.037}  
		& \textbf{0.073} / 0.046                        \\ \cline{2-7} \cline{9-10} 
		\textbf{traffic-1d} & 0.223 / 0.137    & 0.236 / 0.148  & 0.186 / -       & 0.167 / 0.113   & 0.159 / 0.106     & 0.152 / 0.102        &  & 0.123 / 0.099                     & \textbf{0.120} / \textbf{0.083}                        \\
		\textbf{traffic-7d} & 0.492 / 0.280    & 0.509 / 0.529  & 0.202 / -       & 0.168 / 0.114   & 0.251 / 0.169     & 0.172 / 0.110        &  & 0.246 / 0.167                     & \textbf{0.155} / \textbf{0.098}                        \\ \thickhline
	\end{tabular}
	
\end{table*}

\begin{table*}[]
	\fontsize{7.5}{12}\selectfont
	\centering
	\caption{Evaluations summary, using $QL_{\rho=0.5}$/$QL_{\rho=0.9}$ metrics, on Solar, M4 and Wind datasets with different forecast windows, where \textsuperscript{$\triangle$} are extracted from \cite{li2019enhancing}.}
	\label{table:real_datasets_2}
	\begin{tabular}{cccclcc}
		\thickhline
		\textbf{Dataset} & \multicolumn{6}{c}{\textbf{Method}}                                                                                            \\ \thickhline
		\textbf{}        & \textbf{TRMF\textsuperscript{$\triangle$}} & \textbf{DeepAR} & \textbf{ConvTrans} &  & \textbf{SAAM (DeepAR)} & \textbf{SAAM (ConvTrans)} \\ \cline{2-4} \cline{6-7} 
		\textbf{Solar}   & 0.241 / -     & 0.222 / 0.093   & 0.210 / 0.082      &  & \textbf{0.191} / \textbf{0.066}                   & 0.197 / 0.069                      \\
		\textbf{M4}      & - / -         & 0.085 / 0.044   & 0.067 / 0.049      &  & \textbf{0.048} / \textbf{0.029}                   & 0.061 / 0.044                      \\
		\textbf{Wind}    & 0.311 / -     & 0.286 / 0.116   & 0.287 / 0.111      &  &  0.282 /  \textbf{0.105}                   & \textbf{0.278} /0.108                      \\ \thickhline
	\end{tabular}
\end{table*}

\begin{table*}[]
	\fontsize{7.5}{12.5}\selectfont
	\centering
	\caption{Improvement percentage on $QL_{\rho=0.5}$/$QL_{\rho=0.9}$ metrics for each base model.}
	\label{table:real_datasets_comparative}
	\begin{tabular}{ccccccccc}
		\thickhline
		\multicolumn{2}{c}{\textbf{Dataset}}                 & \textbf{elect-1d} & \textbf{elect-7d} & \textbf{traffic-1d} & \textbf{traffic-7d} & \textbf{Solar} & \textbf{M4}   & \textbf{Wind} \\\thickhline
		\multirow{2}{*}{\textbf{Base Model}} & \textbf{DeepAR}    & 15.33\% / 20.75\%     & 7.32\% / 30.19\%      & 22.64\% / 6.60\%        &1.99\% / 1.18\%        & 13.96\% / 29.03\%  & 43.53\% / 34.09\% & 1.05\% / 9.48\%   \\ \cline{2-9} 
		& \textbf{ConvTrans} & 5.08\% / 14.71\%      & 7.59\% / 9.80\%       & 21.1\% / 18.6\%         & 9.8\% / 10.91 \%       & 6.19\% / 15.85\%   & 8.96\% / 10.20\%  & 3.14\% / 2.70\%   \\ \thickhline
	\end{tabular}
	
\end{table*}

	
	\hfil

	 Table \ref{table:real_datasets} shows the results obtained for both electricity and traffic datasets, with forecast windows of 1 and 7 days. Table \ref{table:real_datasets_2} shows the results for solar, M4, and wind datasets.

	Some conclusions can be drawn from these results. The classic methods evaluated, ARIMA and ETS, performed the worst, probably due to the incapacity of detecting shared patterns across the different time series. The results reported for TRMF are slightly better but, for most configurations, it was not capable to beat Deep Neural Networks based approaches, where DeepAR and ConvTrans solidly exceeded DSSM. The two proposed variations of SAAM outperformed all other  models, as it happened in Section \ref{synthetic} with the synthetic dataset experiments.
	
	Finally, Table \ref{table:real_datasets_comparative} shows a comparison between the base models, DeepAR and ConvTrans, and their SAAM version, which consistently improved base models' forecast accuracy.
	
	For DeepAR, $QL_{\rho=0.5}$ was improved by a 15.1\% and $QL_{\rho=0.9}$ by a 18.8\%  after including the SA module on SAAM. For ConvTrans, $QL_{\rho=0.5}$ was improved by a 8.8\% and  $QL_{\rho=0.9}$ by a 11.8\% when using our proposed SAAM architecture.
	
	These results prove how different deep-autoregressive mo\-dels, with significant differences between them, can be improved by correctly incorporating frequency domain in\-for\-ma\-tion into the forecasting, without any significant complexity overload.
	
	\subsection{Ablation study}
	\label{ablation}

	Finally, to quantify the real effect of the SA module and understand the effectiveness of its components, an ablation study was conducted. SA's basic blocks: the Local Spectral Attention model and the Global Spectral Attention model, described in Section \ref{FAAM}, were separately evaluated.
	
	To evaluate the behavior of both frequency-domain attention mo\-dels, two ablation studies with two different datasets were conducted to secure the robustness of the conclusions. For both studies, SAAM is trained using a LSTM as em\-be\-dding function. 
	
	Considering that SA's output obeys to: $ \mathbf{A}_{t}^{i}=\mathbf{L}_{t}^{i} \odot  \boldsymbol{\alpha}_{l, t}^{i} +\mathbf{G}_{t}^{i} \odot  \boldsymbol{\alpha}_{g, t}^{i}, \in \mathbb{R}^{D \times N_{\mathcal{F} \mathcal{T}}} $, as stated in Table (\ref{eq:fam}), three different configurations of the model were tested: 1) SAAM; 2) SAAM without using Local Spectral Attention, $\boldsymbol{\alpha}_{l, t}^{i} = 1$; 3) SAAM without using Global Spectral Attention, $\boldsymbol{\alpha}_{g, t}^{i} = 0$.

	\begin{table*}[]
		\fontsize{7.5}{12.5}\selectfont
		\centering
		\caption{Ablation study on the synthetic dataset.}
		\label{table:ablation_study_synthetic}
		\begin{tabular}{ccllll}
			\thickhline
			\textbf{}                       & \textbf{}          & \multicolumn{1}{c}{\textbf{ND}}                  & \multicolumn{1}{c}{\textbf{RMSE}}                & \multicolumn{1}{c}{ \textbf{$\mathbf{QL_{\rho=0.5}}$}}               & \multicolumn{1}{c}{ \textbf{$\mathbf{QL_{\rho=0.9}}$}}               \\\thickhline
			& \textbf{Full SAAM}    & \textbf{0.028 $\pm$ 0.000} & \textbf{0.044 $\pm$ 0.000} & \textbf{0.029 $\pm$ 0.000} & \textbf{0.025 $\pm$ 0.000} \\
			& \textbf{No global SA} & 0.886 $\pm$ 0.003          & 1.064 $\pm$ 0.004          & 0.893 $\pm$ 0.003          & 1.077 $\pm$ 0.009          \\
			\multirow{-3}{*}{ \textbf{$\sigma^{2}_{\nu}=0$} }   & \textbf{No local SA}  & 2.084$\pm$ 0.000           & 2.352 $\pm$ 0.000          & 2.100 $\pm$ 0.000          & 2.416 $\pm$ 0.001          \\ \hline
			& \textbf{Full SAAM}    & \textbf{0.816 $\pm$ 0.000} & \textbf{1.084 $\pm$ 0.000} & \textbf{0.823 $\pm$ 0.001} & \textbf{0.859 $\pm$ 0.001} \\
			&\textbf{No global SA}& 0.949 $\pm$ 0.007          & 1.167$\pm$ 0.009           & 0.956 $\pm$ 0.007          & 1.136 $\pm$ 0.007          \\
			\multirow{-3}{*}{ \textbf{$\sigma^{2}_{\nu}=0.5$} } & \textbf{No local SA} & 1.936 $\pm$ 0.000          & 2.263 $\pm$ 0.000          & 1.962 $\pm$ 0.000          & 1.936 $\pm$ 0.001          \\ \hline
			& \textbf{Full SAAM}    & \textbf{0.911 $\pm$ 0.000} & \textbf{1.145 $\pm$ 0.000} & \textbf{0.915 $\pm$ 0.000} & \textbf{0.891 $\pm$ 0.001} \\
			& \textbf{No global SA}& 0.990 $\pm$ 0.006          & 1.231 $\pm$ 0.005          & 0.995 $\pm$ 0.006          & 1.126 $\pm$ 0.008          \\
			\multirow{-3}{*}{ \textbf{$\sigma^{2}_{\nu}=1$} }   & \textbf{No local SA}  & 1.567 $\pm$ 0.000          & 1.890 $\pm$ 0.000          & 1.591 $\pm$ 0.001          & 1.493 $\pm$ 0.002          \\ \thickhline
		\end{tabular}
	\end{table*}
	
	\begin{table*}[]
		\fontsize{7.5}{12.5}\selectfont
		\centering
		\caption{Ablation study on the electricity dataset with 7 days forecast windows.}
		\label{table:ablation_study_elect}
		\begin{tabular}{ccccc}
			\thickhline
			\textbf{}          & \textbf{ND}              & \textbf{RMSE}            & \textbf{$\mathbf{QL_{\rho=0.5}}$}         & \textbf{$\mathbf{QL_{\rho=0.9}}$}             \\\thickhline
			\textbf{Full SAAM}    & \textbf{0.076 $\pm$ 0.000} & \textbf{0.496 $\pm$ 0.001} & \textbf{0.076 $\pm$ 0.000} & \textbf{0.03759 $\pm$ 0.000} \\
			\textbf{No global SA} & 0.236 $\pm$ 0.001          & 2.283 $\pm$ 0.029          & 0.237 $\pm$ 0.001          & 0.078 $\pm$ 0.001            \\
			\textbf{No local SA}  & 0.263 $\pm$ 0.000          & 2.665 $\pm$ 0.001          & 0.264 $\pm$ 0.000          & 0.092 $\pm$ 0.000            \\ \thickhline
		\end{tabular}
	\end{table*}

	Note that fixing $\boldsymbol{\alpha}_{l, t}^{i} = 1$ makes no change in the local context, which implies that no filtering is performed. Besides, $\boldsymbol{\alpha}_{l, t}^{i} = 0$ disables models' habilities to incorporate global trends into the local context.
	
	The first ablation study used the synthetic dataset explained in Section \ref{synthetic} with a noise component of $\sigma^{2}_{\nu}=0$, no covariates and a forecast window of $\{t_0 = 120, \tau = 80\}$. Table \ref{table:ablation_study_synthetic} shows the obtained results. The degradation produced by the ablation when $\sigma^{2}_{\nu}=0$  is bigger than other considered cases, which is normal considering that the model was trained on those conditions.	Also, to disable the local attention $\boldsymbol{\alpha}_{l, t}^{i} = 1$, produced worse results than $\boldsymbol{\alpha}_{g, t}^{i} = 0$, when the global attention is not used. The later causes a bigger deterioration in predictions' variance, which could be a sign of the model's inability to follow the trend after removing the global attention model.

	A second ablation study was performed on the electricity dataset using a 7 days forecasting window. Again, the Local and  Global Spectral Attention models are separately evaluated. For this dataset, the degradation of the results is similar for both ablations, as Table \ref{table:ablation_study_elect} shows. As in the synthetic dataset ablation study, not using  the Global Spectral Attention model translates into a higher standard deviation on the results.

	\section{Conclusion}
	\label{conclusion}
	
	We have proposed a novel methodology for neural pro\-ba\-bi\-lis\-tic time series forecasting that marries signal processing techniques with deep learning-based autoregressive models, developing an attention mechanism which operates over the frequency domain. Thanks to this combination, which is enclosed in the Spectral Attention module, local spectrum filtering and global patterns incorporation meet during the forecast. To do so, two attention models operate over the embedding's spectral domain representation to determine, at every time instant and for each time series, which components of the frequency domain should be considered noise and hence be filtered out, and which global patterns are relevant and should be incorporated into the predictions. Experiments on synthetic and real-world datasets confirm these statements and unveil how our suggested modular architecture can be incorporated into a variety of base deep-autoregressive models, consistently improving the results of these base models and achieving state-of-the-art performance. Especially, in noisy environments or short conditioning ranges, our method stands out by explicitly filtering the noise and rapidly recognizing relevant trends.

	\FloatBarrier
	
	

	\section*{Acknowledgment}
	
	This work has been supported by Spanish government Ministerio de Ciencia, Innovación y Universidades under grants FPU18/00470, TEC2017-92552-EXP and RTI2018-099655-B-100, by Comunidad de Madrid under grants IND2017/TIC-7618, IND2018/TIC-9649, IND2020/TIC-17372,  and Y2018/TCS-4705, by BBVA Foundation under the Deep-DARWiN project, and by the European Union (FEDER) and the European Research Council (ERC) through the European Union’s Horizon 2020 research and innovation program under Grant 714161.
	
	\ifCLASSOPTIONcaptionsoff
	\newpage
	\fi

	
	
	%
	
	\bibliographystyle{IEEEtran}
	\bibliography{IEEEexample}

	%
	
	
	
	
	
	
	

\end{document}